\documentclass[10pt,conference]{IEEEtran}

\usepackage{cite}
\usepackage{comment}
\ifCLASSINFOpdf
   \usepackage[pdftex]{graphicx}
   \graphicspath{{figs/}}
   \DeclareGraphicsExtensions{.pdf,.jpeg,.png}
\else
   \usepackage[dvips]{graphicx}
   \graphicspath{{../figs/}}
   \DeclareGraphicsExtensions{.eps}
\fi
\usepackage{multirow}

\usepackage[cmex10]{amsmath}
\usepackage{amsthm}

\usepackage{algorithmic}
\usepackage{array}
\usepackage{xcolor}

\ifCLASSOPTIONcompsoc
  \usepackage[caption=false,font=normalsize,labelfont=sf,textfont=sf]{subfig}
\else
  \usepackage[caption=false,font=footnotesize]{subfig}
\fi
\usepackage{url}
\newif\iffinal
\finaltrue
\newcommand{\cmtid}{99999}
\iffinal
\else
\usepackage[switch]{lineno}
\fi

\begin{document}
\title{Emotion Intensity Matters: Generating Realistic Expressions in Virtual Humans with CVAEs}


\iffinal

\author{
\IEEEauthorblockN{Vitor Miguel Xavier Peres, Lara Volpato, Gabriel Scnheider and Soraia Raupp Musse}
\IEEEauthorblockA{School of Technology\\Pontifical Catholic University of Rio Grande do Sul\\
Porto Alegre, Brazil\\
Email: \{vitor.peres,lara.volpato,gabriel.ferri\}@edu.pucrs.br and soraia.musse@pucrs.br}
% \and
% \IEEEauthorblockN{Soraia Raupp Musse}
% \IEEEauthorblockA{School of Technology\\Pontifical Catholic University of Rio Grande do Sul\\
% Porto Alegre, Brazil\\
% Email: soraia.musse@pucrs.br}
% \and
% \IEEEauthorblockN{James Kirk\\ and Montgomery Scott}
% \IEEEauthorblockA{Starfleet Academy\\
% San Francisco, California 96678--2391\\
% Telephone: (800) 555--1212\\
% Fax: (888) 555--1212}
}


%Telephone: (800) 555--1212, Fax: (888) 555--1212}

\else
  \author{SIBGRAPI Paper ID: \cmtid \\ }
  \linenumbers
\fi

\maketitle

\begin{abstract}

Generating expressive facial behavior in virtual humans (VHs) remains a central challenge in affective computing and character animation. This paper presents a novel approach based on Conditional Variational Autoencoders (CVAEs), trained on real human facial expression data, to synthesize controllable emotional expressions at varying intensities. Using a dataset comprising six basic emotions represented at two intensity levels (low and high), we train a CVAE model to generate synthetic facial expression data while preserving semantic consistency with real human expressions. Despite the limited amount of training data (only 7,680 facial expression samples), the proposed approach learns meaningful latent representations and generates coherent emotional variations. Our method enables control over emotional intensity, making it suitable for animating virtual characters without requiring actor performances or manual artistic intervention. Our research aimed to evaluate whether the method (CVAE) preserves the characteristics associated with the different intensity levels present in the dataset. Results show that the proposed model preserves key expressive characteristics across intensity levels while supporting generalization across emotional intensity levels, contributing to the creation of emotionally expressive virtual characters from relatively small datasets.

\end{abstract}


\IEEEpeerreviewmaketitle

\section{Introduction}

Virtual humans (VHs) are increasingly present in interactive environments, ranging from games and simulations to healthcare, education, and social interaction.
As their presence expands, so does the demand for emotionally expressive and believable characters capable of nuanced affective behavior. Among the many challenges in this domain, generating naturalistic and contextually appropriate facial expressions stands out as a core research problem, for which, in general, the solution involves artistic and performance-based techniques. Since the seminal work of Lance~\cite{Lance1990} introducing Performance-Driven Animation (PDA), computer vision techniques have been used to capture facial expressions and apply them to virtual faces. 
While PDA enables the transfer of real human expressions to virtual faces, there is also growing interest in generating virtual humans that can imitate human expressions, without requiring expression transfer from a real person or artist. 

One of the seminal contributions to this area is the morphable model proposed by Blanz and Vetter~\cite{Blanz1999}. Their approach transforms a dataset of textured 3D face scans into a high-dimensional linear model, enabling the synthesis and manipulation of realistic 3D faces through linear combinations of prototype faces. This work established a statistically grounded framework for facial analysis and synthesis and introduced the concept of a face space that continues to influence modern facial modeling techniques. Despite the success of morphable models in representing facial identity and appearance, generating expressive facial behavior remains a considerably more challenging problem. Unlike facial shape, emotional expressions are inherently dynamic and vary in intensity, making them difficult to model and reproduce in virtual humans without relying on actor performances or manual artistic authoring. This challenge becomes even more pronounced when only limited training data are available, as collecting and annotating high-quality facial expression datasets is both expensive and time-consuming.

To address this problem, we investigate the use of Conditional Variational Autoencoders (CVAEs) to generate emotional expressions for VHs from a relatively small set of real facial expression samples. The model is trained on six basic emotions (happiness, anger, fear, sadness, disgust, and surprise) represented at two intensity levels (low and high). By conditioning the generation process on emotion, intensity, and gender labels, the CVAE enables controllable facial expression synthesis while preserving semantic consistency across emotional states.

Our method addresses three key challenges:
\begin{itemize}
\item Learning expressive facial representations from limited real-world training data;
\item Generating controllable emotional expressions across multiple intensity levels without artistic authoring or actor-specific supervision;
\item Refining CVAE-generated animations through a residual enhancement framework that recovers facial dynamics and emotional intensity using temporal residual learning.
\end{itemize}

\section{Related Work}

Autoencoders are neural network architectures consisting of three interconnected components: an encoder, a latent space, and a decoder. The encoder reduces the input data to a lower-dimensional latent space, while the decoder attempts to reconstruct the original input data from this latent representation. The Variational Autoencoder (VAE) employs a structured and continuous latent space as its fundamental learning framework~\cite{ISLAM2021105950}. This approach facilitates greater control and more comprehensive exploration during data generation, thereby enabling the synthesis of diverse, variable datasets. Furthermore, the VAE effectively projects data into the latent space, ensuring smooth transitions within the representation. In practice, this feature is particularly beneficial for distinguishing subtle variations, such as differences in facial expression intensity. Additionally, the probabilistic nature of the VAE enhances its ability to generalize to unseen data, underscoring its relevance for tasks that require generating diverse and representative outputs.

%Most similar to our work, Qiu et al.~\cite{10.1145/3680528.3687669} propose FreeAvatar, a 3D facial animation transfer method based solely on expression representation, by utilizing a mixed Masked Autoencoder (MAE)~\cite{he2022masked} and GAN architecture. 
%First, an expression feature space is built by training the MAE on 4.5 million unlabeled facial images and then refined by fine-tuning the encoder on 914 thousand labeled expression triplet sets. Following that, a rig parameter decoder and a neural renderer are used to enable identity injection, allowing the expression to be rendered on any avatar, provided the rig parameter decoder was trained on that avatar. The user study suggests better transfer than other state-of-the-art methods.

Sohn et al.~\cite{NIPS2015_8d55a249}, introduces the Conditional Variational Autoencoder (CVAE) model, a probabilistic extension of the Variational Autoencoder (VAE) applied to structured prediction problems. The work presents a deep generative architecture capable of modeling complex, multimodal conditional distributions, enabling the generation of multiple plausible outputs from the same input. While prior works such as FreeAvatar~\cite{10.1145/3680528.3687669}, EVAE~\cite{Bai:2022}, and VAE-GAN~\cite{YAN2025112899} have demonstrated promising applications of generative models in expression transfer, arousal-valence modeling, or privacy-preserving synthesis, they often rely on external identity injection or fixed semantic spaces. Our approach differs by introducing intensity-specific CVAEs trained on real-world expressive data and transferring this expressiveness to virtual agents without requiring actor-specific supervision or preconditioning. 

Regarding Residual Networks (ResNets), He et al.~\cite{he2015deepresiduallearningimage} introduced the concept of residual learning to address the degradation problem commonly observed in very deep neural networks. Instead of learning a direct mapping between inputs and outputs, the proposed architecture learns residual functions relative to the input, facilitating optimization and improving convergence. The authors demonstrated significant performance gains in image recognition tasks, achieving state-of-the-art results on ImageNet and COCO benchmarks. Beyond image classification, the residual learning paradigm has inspired numerous applications aimed at modeling reconstruction errors and recovering information lost during data generation processes. In this work, we adopt a similar principle by computing the residual difference between real facial expression data and the corresponding expressions generated by the CVAE. These residual sequences are subsequently used to train a Long Short-Term Memory (LSTM) network, enabling the modeling of temporal patterns associated with the reconstruction errors. The residuals predicted by the LSTM are then applied to the original CVAE-generated data, enhancing facial dynamics, restoring emotional intensity, and improving the overall realism of the synthesized expressions.

\section{Methodology}

\begin{figure*}[t!]
    \centering
    \includegraphics[width=16cm]{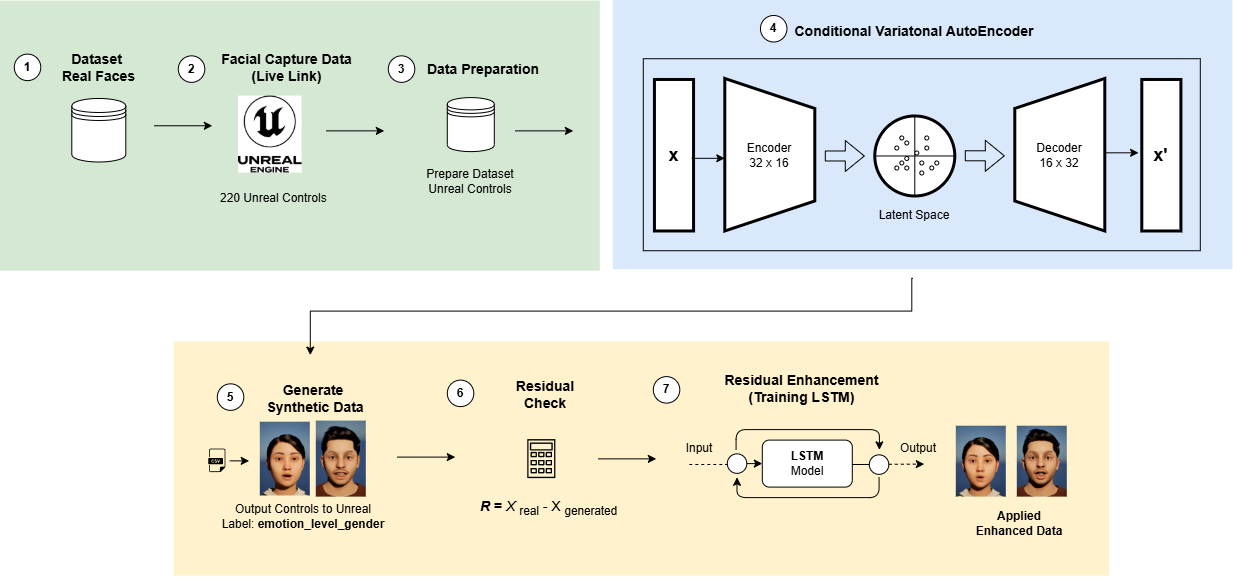}
    \caption{Method Overview: (1) Obtaining dataset; (2) Preparing to Capture Data (Live Link Hub); (3) Transferring the Capture Data to Metahuman and Prepare a dataset to (4) Training a CVAE and (5) the Generation of new facial expressions using the CVAE output, for each level of emotion intensity; Residual Check (6) from Generated Data and Residual Enhancement (7).}
    \label{fig:method}
\end{figure*}

The overview of our methodology is illustrated in Figure~\ref{fig:method}. In Step 1, we collected the MEAD~\cite{kaisiyuan2020mead} and RAVDESS~\cite{livingstone_2018_1188976} datasets, comprising 864 videos of individuals performing the six basic emotions at two intensity levels (Low and High). In Step 2, we used the Live Link Hub\footnote{https://dev.epicgames.com/documentation/unreal-engine/live-link-hub-quick-start-in-unreal-engine} plugin in Unreal Engine to process each video and generate Facial Capture Data. This data contains frame-by-frame facial expression information, including head pose, facial landmark tracking, and the animation curves used to drive the MetaHuman facial rig. From these sequences, we extracted the MetaHuman facial parameters, which correspond to the values of the facial controls that drive the character's expressions over time, thereby providing a compact numerical representation of the emotion performed. The extracted controls include facial movements of the eyes, eyebrows, eyelids, nose, cheeks, jaw, and mouth, as well as head motion parameters, yielding a temporal representation of each facial performance that can serve as input for subsequent modeling and expression synthesis.

These parameters were compiled to form a CVAE Training Dataset (Step 3) and used in Step 4 to train the Variational Autoencoder (CVAE). The trained CVAE then generates new MetaHuman parameters representing novel facial expressions based on input information corresponding to the training classes, e.g., women/men, happy (emotions), and low/high intensity. These outputs are directly applied to a MetaHuman model to synthesize facial expressions (Step 5). 

We also propose a residual enhancement stage in which the differences between the real and CVAE-generated facial expressions (Step 6) are modeled using a Long Short-Term Memory (LSTM) network (Step 7). By learning the structure of these residual signals, the LSTM predicts corrective facial dynamics that are subsequently applied to the original synthetic data. Further details about all steps explicit in Figure~\ref{fig:method} are described in the following sections.

\subsection{Dataset Preprocessing (1)}
\label{sec:1_e_4}

In this paper, we used the MEAD (Multi-view Emotional Audio-visual Dataset)~\cite{kaisiyuan2020mead} and RAVDESS (The Ryerson Audio-Visual Database of Emotional Speech and Song)~\cite{livingstone_2018_1188976} datasets, which feature actors and actresses talking while expressing six different emotions at three different intensity levels (Low, Medium, and High). High-quality audio-visual clips are captured in a strictly controlled environment at seven different angles. Still, in the present paper, we only used front-view capture and two intensity levels (low and high) to test more distinct intensities that represent the extremes of facial expression variation in the dataset. In total, we randomly selected 16 people, 8 men and 8 women, and for each emotion we selected 2 videos, each video showing two different intensities. This resulted in a total of 384 videos. 

\subsection{Capturing (2) and Preparing (3) Data} 

The Live Link Hub plugin (in Unreal Engine) is used to generate, by processing videos, the Facial Capture Data, which contains frame-by-frame facial expression information. From these sequences, we extracted the MetaHuman facial parameters, which correspond to the values of the facial controls that drive the character's expressions over time, thereby providing a compact numerical representation of the emotion performed. Then, we exported the controls from the Face\_ControlBoard\_CTRLRig to a CSV file. This file contains 220 facial controls (attributes) and a sequence of rows corresponding to the facial motion timeline, with each row representing a frame of the animation. Together, these controls and temporal samples constitute the dataset used to train the CVAE. Since the videos contain individuals speaking while displaying facial expressions, many frames are influenced by speech articulation and therefore represent visemes rather than the intended emotion. To reduce this effect, we performed a frame selection procedure to identify the frames that best represent each emotional expression. We computed an intensity score based on the absolute activation of the MetaHuman control parameters. For each frame, the score was calculated as $S = \sum_{i=1}^{220} |C_i|,$ where $C_i$ represents the value of the $i$-th facial control parameter. Frames with higher values of $S$ correspond to stronger facial activations and, consequently, more expressive emotional states. For each video, we retained the 20 frames with the highest scores. Considering the 384 videos in the dataset (16 subjects $\times$ 6 emotions $\times$ 2 intensity levels $\times$ 2 recordings per subject), this procedure yielded 7,680 frames for subsequent analysis.

\subsection{CVAE Training (4)}

We adopt a Conditional Variational Autoencoder (CVAE) rather than a standard VAE because the generation process must be explicitly guided by semantic attributes. The conditional formulation allows the latent representation to be associated with emotion category, intensity level, and gender labels, enabling controllable facial expression synthesis rather than unconstrained sample generation. Our CVAE architecture consists of a two-layer fully connected encoder followed by a decoder with two hidden layers. The encoder contains 32 and 16 neurons, respectively, and maps the 220-dimensional Unreal facial control representation into a latent space. The decoder receives a latent-space vector and reconstructs the original 220 facial control values. It consists of two fully connected hidden layers with 16 and 32 neurons, respectively, followed by an output layer with 220 neurons. A sigmoid activation function is applied to the output layer to match the normalized range of the facial control parameters.

The model was trained using the Adam optimizer, with a learning rate of 0.0001 and a batch size of 32, over 400 epochs. The training process was developed in Python using the Keras library and the TensorFlow backend. It was conducted on a PC running Windows 11 Pro, an Intel Core i9 14900KF at 5.8 GHz, 64GB of RAM at 5200 MT/s, a 1 TB NVME SSD, and an RTX 3070 GPU with 8 GB of memory. Training took an average of 18 minutes, 44.7s per emotion (low and high intensities), and 2.122s per epoch.

\subsection{Generated Synthetic Data (5)}

Figure~\ref{fig_sim} presents an example of a subject whose 220 facial control parameters were used to train the CVAE. On the left, is a real frame from one of the videos. To generate facial expressions, the CVAE, central image of Figure~\ref{fig:samples}, is conditioned on a multi-label vector containing the target emotion, gender, intensity level, and optionally the video identifier, which serves as a proxy for the subject's identity. 

%\begin{figure}[h]
%    \centering
%    \includegraphics[width=.70\linewidth]{figs/imgpaper.png}
%    \caption{Sample of a real and transferred facial expression. We also present results from applying CVAE and an enhanced CVAE for the emotion of Anger.}
%    \label{fig:samples}
%\end{figure}

\subsection{Residual Check (6) and Enhancement (7)}

To reduce the smoothing of the synthetic data generated by the Conditional Variational Autoencoder (CVAE), an additional enhancement step based on residual learning was employed. Initially, the residual between the real and synthetic data generated by the CVAE was calculated as the difference between the corresponding facial control values in each frame. Then, a Long Short-Term Memory (LSTM) network was trained to model this residual, exploiting its ability to capture dynamic patterns present in facial expressions. This approach is inspired by the principles of residual learning introduced by He~\cite{he2015deepresiduallearningimage}, in which the network directly learns the correction function rather than the complete reconstruction of the signal, thereby reducing the complexity of the learning problem. During training, residual sequences are computed as the frame-wise differences between the real facial-control sequences and their corresponding CVAE reconstructions. The LSTM is trained to model the structure of these residual sequences. In the current enhancement procedure, the residual derived from the paired real and CVAE-generated sequences is provided to the LSTM, whose refined output is then added to the CVAE generation.
%During training, the network learns the relationship between the CVAE output and the corresponding real frame by analyzing the residual between them. In other words, the model learns a correction pattern representing how the synthesized data can be adjusted to approximate the distribution of the real data, without requiring access to the corresponding real frame.} 
%After training, the LSTM began predicting the missing residual components in the synthesized data, which were then added to the sequences originally generated by the CVAE. 

Figure~\ref{fig_sim}, on the right, presents an example generated by the CVAE Enhanced model, which exhibits more pronounced mouth deformations than the standard CVAE (in the center). It is worth noting that neither model performs direct expression transfer from the real footage shown on the left. Instead, both learn a latent representation of facial behavior from the training data and generate new expressions conditioned on the input labels. As a result, emotional facial animations can be synthesized without the need for actor capture or manual artistic editing.

\section{Experimental Results}

\begin{table}[t]
\centering
\tiny
\setlength{\tabcolsep}{3pt}
\resizebox{\columnwidth}{!}{
\begin{tabular}{c|cc|cc|cc|cc}

\textbf{} &
    \multicolumn{4}{c|}{\textbf{Men}} &
    \multicolumn{4}{c}{\textbf{Women}} \\ \hline
    
    \textbf{} &
    \multicolumn{2}{c}{\textbf{Low}} &
    \multicolumn{2}{c|}{\textbf{High}} &
    \multicolumn{2}{c}{\textbf{Low}} &
    \multicolumn{2}{c}{\textbf{High}} \\ \hline
    
    \textbf{Emotion} &
    \textbf{\textit{p}} & \textbf{$\sigma$} &
    \textbf{\textit{p}} & \textbf{$\sigma$} &
    \textbf{\textit{p}} & \textbf{$\sigma$} &
    \textbf{\textit{p}} & \textbf{$\sigma$} \\ \hline
    
    \textbf{Angry}     & 0.370 & 0.320 & 0.364 & 0.311 & 0.302 & 0.323 & 0.308 & 0.318 \\ 
    \textbf{Disgusted} & 0.386 & 0.318 & 0.353 & 0.321 & 0.278 & 0.315 & 0.275 & 0.314 \\ 
    \textbf{Fear}      & 0.367 & 0.326 & 0.360 & 0.310 & 0.282 & 0.308 & 0.313 & 0.314 \\ 
    \textbf{Happy}     & 0.361 & 0.316 & 0.296 & 0.325 & 0.320 & 0.312 & 0.308 & 0.318 \\ 
    \textbf{Sad}       & 0.400 & 0.309 & 0.338 & 0.322 & 0.339 & 0.318 & 0.319 & 0.315 \\ 
    \textbf{Surprised} & 0.357 & 0.315 & 0.349 & 0.309 & 0.295 & 0.308 & 0.275 & 0.315 \\ \hline    
    \end{tabular}
}
\caption{Average p-values and standard deviations between \textbf{real} and \textbf{synthetic} data, for each emotion according to gender and intensity level.}
\label{tab1}
\end{table}

\begin{table}[t]
\centering
\tiny
\setlength{\tabcolsep}{3pt}
\resizebox{\columnwidth}{!}{
\begin{tabular}{c|cc|cc|cc|cc}

\textbf{} &
    \multicolumn{4}{c|}{\textbf{Men}} &
    \multicolumn{4}{c}{\textbf{Women}} \\ \hline
    
    \textbf{} &
    \multicolumn{2}{c}{\textbf{Low}} &
    \multicolumn{2}{c|}{\textbf{High}} &
    \multicolumn{2}{c}{\textbf{Low}} &
    \multicolumn{2}{c}{\textbf{High}} \\ \hline
    
    \textbf{Emotion} &
    \textbf{\textit{p}} & \textbf{$\sigma$} &
    \textbf{\textit{p}} & \textbf{$\sigma$} &
    \textbf{\textit{p}} & \textbf{$\sigma$} &
    \textbf{\textit{p}} & \textbf{$\sigma$} \\ \hline
    
    \textbf{Angry}     & 0.648 & 0.318 & 0.591 & 0.323 & 0.616 & 0.329 & 0.631 & 0.322 \\ 
    \textbf{Disgusted} & 0.606 & 0.314 & 0.594 & 0.320 & 0.599 & 0.322 & 0.597 & 0.317 \\ 
    \textbf{Fear}      & 0.614 & 0.321 & 0.657 & 0.318 & 0.602 & 0.315 & 0.635 & 0.315 \\ 
    \textbf{Happy}     & 0.698 & 0.319 & 0.670 & 0.317 & 0.615 & 0.314 & 0.597 & 0.324 \\ 
    \textbf{Sad}       & 0.615 & 0.313 & 0.568 & 0.324 & 0.565 & 0.314 & 0.552 & 0.318 \\ 
    \textbf{Surprised} & 0.640 & 0.311 & 0.650 & 0.306 & 0.581 & 0.310 & 0.595 & 0.317 \\ \hline  
    \end{tabular}
}
\caption{Average p-values and standard deviations between \textbf{real} and \textbf{enhanced} synthetic data, for each emotion according to gender and intensity level.}
\label{tab2}
\end{table}

\begin{table}[t]
\centering
\tiny
\setlength{\tabcolsep}{3pt}
\resizebox{\columnwidth}{!}{
\begin{tabular}{c|cc|cc|cc|cc}

\textbf{} &
    \multicolumn{4}{c|}{\textbf{Men}} &
    \multicolumn{4}{c}{\textbf{Women}} \\ \hline
    
    \textbf{} &
    \multicolumn{2}{c}{\textbf{Low}} &
    \multicolumn{2}{c|}{\textbf{High}} &
    \multicolumn{2}{c}{\textbf{Low}} &
    \multicolumn{2}{c}{\textbf{High}} \\ \hline
    
    \textbf{Emotion} &
    \textbf{\textit{p}} & \textbf{$\sigma$} &
    \textbf{\textit{p}} & \textbf{$\sigma$} &
    \textbf{\textit{p}} & \textbf{$\sigma$} &
    \textbf{\textit{p}} & \textbf{$\sigma$} \\ \hline
    
    \textbf{Angry}     & 0.655 & 0.265 & 0.758 & 0.233 & 0.632 & 0.244 & 0.633 & 0.241  \\ 
    \textbf{Disgusted} & 0.674 & 0.267 & 0.660 & 0.219 & 0.581 & 0.022 & 0.615 & 0.230  \\ 
    \textbf{Fear}      & 0.653 & 0.241 & 0.646 & 0.253 & 0.721 & 0.242 & 0.686 & 0.236 \\ 
    \textbf{Happy}     & 0.704 & 0.241 & 0.612 & 0.253 & 0.747 & 0.232 & 0.797 & 0.233 \\ 
    \textbf{Sad}       & 0.556 & 0.270 & 0.716 & 0.240 & 0.749 & 0.248 & 0.748 & 0.251 \\ 
    \textbf{Surprised} & 0.673 & 0.241 & 0.601 & 0.233 & 0.707 & 0.210 & 0.832 & 0.175 \\ \hline
    \end{tabular}
}
\caption{Average p-values and standard deviations between \textbf{synthetic} and \textbf{enhanced} data, for each emotion according to gender and intensity level.}
\label{tab3}
\end{table}

\begin{figure*}[!t]
\centering
\includegraphics[width=7.2in]{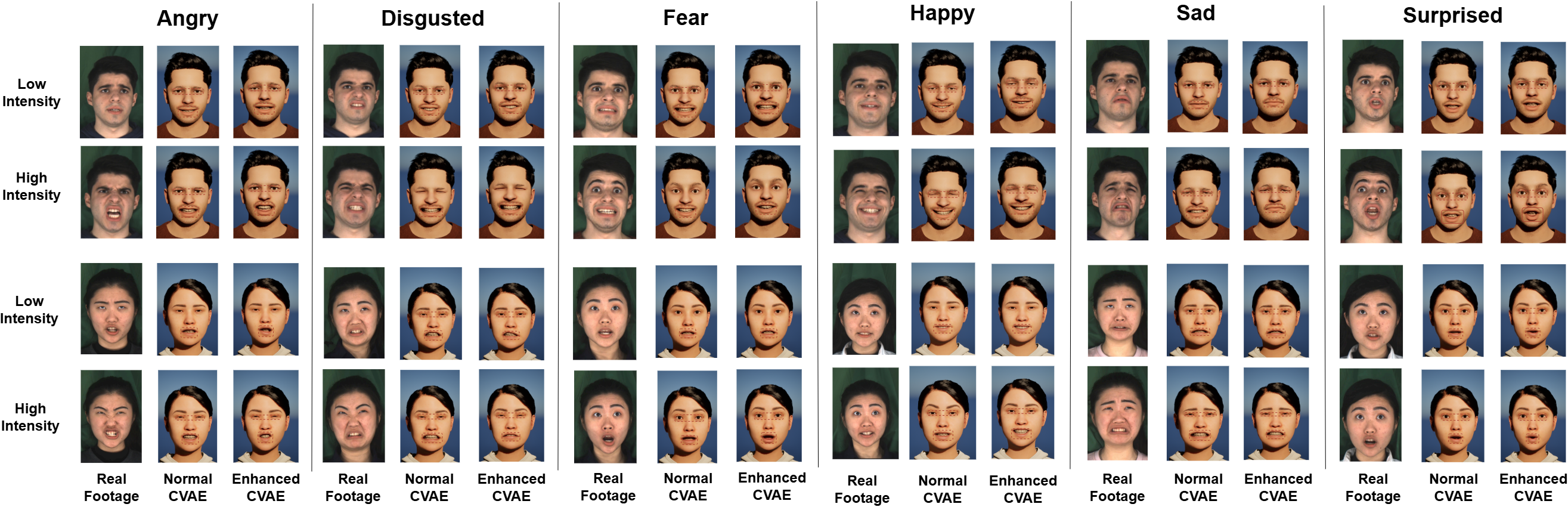}
\caption{Rendering of faces, based on imported data generated via CVAE. We provide a separation into 6 emotions and 2 intensities, and illustrate only 2 individuals. Side-by-side, we show the real face, the face rendered via CVAE, and Enhanced CVAE.}
\label{fig_sim}
\end{figure*}

We performed independent-samples t-tests to compare the Unreal facial control values extracted from the real and synthetic datasets, as detailed in the following sections. Additionally, the generated expressions were evaluated using Py-Feat emotion classification. Qualitative results are illustrated in Figure~\ref{fig_sim}, showing that, although often subtle, the enhancement stage generally amplifies facial dynamics, especially mouth-related expressions. 

\subsection{Comparison Between Real and Generated and Enhanced Generated Data}

Tables~\ref{tab1} and~\ref{tab2} present the equivalence values obtained from statistical comparisons between real and synthetic facial expression data. For each combination of emotion, intensity level, and gender, an independent Student's t-test was performed for each of the 220 Unreal Controls, comparing the corresponding real and synthetic control values across all frames. The reported equivalence value corresponds to the average of the resulting 220 p-values: $\bar{p} = \frac{1}{N}\sum_{i=1}^{N} p_i,$ where \(N = 220\) is the number of Unreal Controls, \(p_i\) is the p-value obtained for control \(i\), and \(\bar{p}\) is the average p-value reported in Tables~\ref{tab1} and~\ref{tab2}. The average p-value is used here as an aggregate indicator of statistical correspondence across all facial controls. Since the null hypothesis of the t-test assumes no difference between the real and synthetic control distributions, larger p-values indicate weaker evidence against this hypothesis and are therefore interpreted as reflecting stronger statistical agreement between the compared datasets. As shown in Tables~\ref{tab1} and~\ref{tab2}, all average p-values are greater than 0.05, indicating no statistically significant differences between the compared datasets and suggesting a strong statistical correspondence between the real and synthetic MetaHuman controls (Table~\ref{tab1}) as well as between the real and enhanced generated data (Table~\ref{tab2}).

In particular, prior to enhancement, the equivalence values ranged from approximately 0.27 to 0.40, indicating a moderate correspondence between the synthetic and real datasets. These results suggest that although the CVAE successfully captured the general structure of facial expressions, it exhibited the smoothing effect commonly associated with variational generative models, attenuating expression peaks and reducing the variability present in the original data. After enhancement, the equivalence values increased consistently across all evaluated conditions, effectively doubling compared to the first generation. This substantial improvement indicates that the residual learning framework effectively corrected the discrepancies between real and generated control trajectories, preserving the expressive dynamics of the original performances while maintaining the controllability provided by the CVAE.

\subsection{Comparison Generated $vs$ Enhanced Generated}

Table~\ref{tab3} presents the equivalence values obtained from the statistical comparison between the synthetic facial expression data generated by the CVAE and the corresponding synthetic data after the residual enhancement stage. The results reveal consistently high equivalence values across all emotions, genders, and intensity levels, ranging from approximately 0.58 to 0.70. An important observation is that the equivalence values between synthetic and enhanced synthetic data are generally higher than those originally observed between real and synthetic data (Table~\ref{tab1}). This finding confirms that the enhancement process operates as a controlled correction mechanism rather than a complete transformation of the generated expressions. The enhanced sequences remain statistically close to the original CVAE output while incorporating additional facial dynamics and expression intensity recovered through temporal residual learning.

Furthermore, the similarity values remain consistently high across both low and high-intensity expressions. This stability suggests that the residual enhancement framework modifies facial activations proportionally rather than indiscriminately increasing expression amplitudes. Consequently, the emotional identity of each expression category is preserved, while intensity-related features are refined and amplified as needed.

% \textbf{Emotion Classification:} To further evaluate the quality of the generated facial expressions, an emotion recognition analysis was conducted using Py-Feat~\cite{cheong2023pyfeatpythonfacialexpression} on the real, CVAE-generated, and Enhanced datasets. For most emotions, recognition performance decreased after CVAE generation but improved substantially following the residual enhancement stage. For example, recognition of angry expressions decreased from 75\% in the real dataset to 50\% in the CVAE-generated data and recovered to 75\% after enhancement. Similar behavior was observed for disgust, which dropped from 100\% to 50\% and improved to 75\%, and for surprised, which recovered from 75\% in the CVAE data to 100\% after enhancement. Fear presented the lowest recognition rates in both the real and CVAE datasets (25\%), likely due to confusion with surprised expressions, but increased markedly to 75\% following enhancement. Sad expressions improved from 75\% recognition in the real dataset to 100\% in both generated versions. Happy expressions showed a different trend, increasing from 50\% in the real dataset to 75\% in the CVAE-generated data, while the Enhanced dataset returned to the original recognition level. Overall, these results suggest that the residual enhancement framework improves the recognizability of generated emotions and, in most cases, brings synthetic expressions closer to the emotional characteristics observed in real data.

\textbf{Emotion Classification}: To further evaluate the quality of the generated facial expressions, an emotion recognition analysis was conducted using Py-Feat~\cite{cheong2023pyfeatpythonfacialexpression} on the real, CVAE-generated, and Enhanced datasets. For most emotions, recognition performance decreased after CVAE generation but improved substantially following the residual enhancement stage.

\begin{itemize}
    \item \textbf{Anger:} The CVAE achieved similar recognition rates for men (56.25\%) and women (50.00\%). The enhanced data improved recognition for men (75.00\%) while maintaining a comparable performance for women (56.25\%).
    \item \textbf{Disgust:} Recognition with the CVAE was moderate for both men (50.00\%) and women (43.75\%). The enhanced data reduced performance for men (31.25\%) but improved recognition for women (62.50\%).
    \item \textbf{Fear:} The CVAE obtained identical recognition rates for men and women (56.25\%). The enhanced data slightly decreased performance for men (50.00\%) and more noticeably for women (37.50\%).
    \item \textbf{Happy:} Happy was the best-recognized emotion in all cases. The CVAE achieved 93.75\% for men and 87.50\% for women, while the enhanced data reached 100.00\% for men and 81.25\% for women.
    \item \textbf{Sad:} Recognition remained relatively low with the CVAE (31.25\% for men and 37.50\% for women). The enhanced data maintained the same performance for men (31.25\%) and improved recognition for women (50.00\%).
    \item \textbf{Surprise:} The CVAE showed lower recognition rates, especially for women (43.75\% for men and 43.75\% for women). The enhanced data substantially improved recognition for men (68.75\%) and moderately increased performance for women (50.00\%).
\end{itemize}

\subsection{Emotions Distances by Intensities Levels}

% \textcolor{blue}{To validate that emotion intensity is preserved in the generated data, we measured the Euclidean distance between the average AU vectors (extracted using OpenFace~\cite{8373812}) among all subjects for each label of low and high expressions for each emotion (Table~\ref{tab:domain_dist}). 

Table ~\ref{tab:domain_dist} summarizes the mean and standard deviation of the estimated emotion intensity for each emotion and target intensity level. As expected, real facial expressions consistently exhibit the highest average intensity, reflecting the greater expressiveness of the original dataset. In contrast, the expressions synthesized by the baseline CVAE exhibit substantially lower intensity across all emotions and levels, indicating that the generated expressions are generally attenuated. The enhanced method consistently increases the average intensity compared with the baseline CVAE while preserving a similar variability, bringing the synthesized expressions closer to the intensity observed in real faces than CVAE. %This trend is particularly evident for higher-intensity expressions (High), where the enhanced model produces a noticeable increase in intensity across all evaluated emotions.

% The Enhanced method increased the separation between intensity levels for four of the six evaluated emotions (angry, fear, sad, and surprised), while disgusted remained nearly unchanged and happy showed a slight reduction.Overall, these findings indicate that the enhancement process produces a clearer distinction between low- and high-intensity expressions, providing quantitative evidence that the generated data encode emotion intensity more effectively than the original CVAE output.}

\begin{table}[h]
    \scriptsize
    \def\arraystretch{1.2}
    \begin{tabular}{clcccccc}
    \hline
\multicolumn{2}{c}{\multirow{2}{*}{\textbf{Emotion}}} & \multicolumn{3}{c}{\textbf{Low}} & \multicolumn{3}{c}{\textbf{High}} \\
\multicolumn{2}{c}{} & \textbf{Real} & \textbf{CVAE} & \textbf{Enh.} & \textbf{Real} & \textbf{CVAE} & \textbf{Enh.} \\ \hline
\textbf{Angry} & \begin{tabular}[c]{@{}l@{}}$\mu$\\ $\sigma$\end{tabular} & \begin{tabular}[c]{@{}c@{}}0.3653\\ 0.5393\end{tabular} & \begin{tabular}[c]{@{}c@{}}0.1857\\ 0.3541\end{tabular} & \begin{tabular}[c]{@{}c@{}}0.2346\\ 0.4274\end{tabular} & \begin{tabular}[c]{@{}c@{}}0.9462\\ 1.2141\end{tabular} & \begin{tabular}[c]{@{}c@{}}0.2906\\ 0.518\end{tabular} & \begin{tabular}[c]{@{}c@{}}0.3418\\ 0.6026\end{tabular} \\ \hline
\textbf{Disgusted} & \begin{tabular}[c]{@{}l@{}}$\mu$\\ $\sigma$\end{tabular} & \begin{tabular}[c]{@{}c@{}}0.6406\\ 0.7553\end{tabular} & \begin{tabular}[c]{@{}c@{}}0.2857\\\ 0.4778\end{tabular} & \begin{tabular}[c]{@{}c@{}}0.3402\\ 0.5388\end{tabular} & \begin{tabular}[c]{@{}c@{}}1.0029\\ 1.3333\end{tabular} & \begin{tabular}[c]{@{}c@{}}0.4297\\ 0.6531\end{tabular} & \begin{tabular}[c]{@{}c@{}}0.4882\\ 0.7363\end{tabular} \\ \hline
\textbf{Fear} & \begin{tabular}[c]{@{}l@{}}$\mu$\\ $\sigma$\end{tabular} & \begin{tabular}[c]{@{}c@{}}0.5818\\ 0.8543\end{tabular} & \begin{tabular}[c]{@{}c@{}}0.2307\\ 0.4449\end{tabular} & \begin{tabular}[c]{@{}c@{}}0.2646\\ 0.4886\end{tabular} & \begin{tabular}[c]{@{}c@{}}0.8494\\ 1.1438\end{tabular} & \begin{tabular}[c]{@{}c@{}}0.3351\\ 0.5949\end{tabular} & \begin{tabular}[c]{@{}c@{}}0.4031\\ 0.6529\end{tabular} \\ \hline
\textbf{Happy} & \begin{tabular}[c]{@{}l@{}}$\mu$\\ $\sigma$\end{tabular} & \begin{tabular}[c]{@{}c@{}}0.3774\\ 0.5524\end{tabular} & \begin{tabular}[c]{@{}c@{}}0.3591\\ 0.5834\end{tabular} & \begin{tabular}[c]{@{}c@{}}0.3644\\ 0.6088\end{tabular} & \begin{tabular}[c]{@{}c@{}}0.9106\\ 1.2511\end{tabular} & \begin{tabular}[c]{@{}c@{}}0.5105\\ 0.8279\end{tabular} & \begin{tabular}[c]{@{}c@{}}0.5353\\ 0.8586\end{tabular} \\ \hline
\textbf{Sad} & \begin{tabular}[c]{@{}l@{}}$\mu$\\ $\sigma$\end{tabular} & \begin{tabular}[c]{@{}c@{}}0.5156\\ 0.7146\end{tabular} & \begin{tabular}[c]{@{}c@{}}0.1883\\ 0.3739\end{tabular} & \begin{tabular}[c]{@{}c@{}}0.1713\\ 0.341\end{tabular} & \begin{tabular}[c]{@{}c@{}}0.8815\\ 0.9397\end{tabular} & \begin{tabular}[c]{@{}c@{}}0.3596\\ 0.5542\end{tabular} & \begin{tabular}[c]{@{}c@{}}0.3722\\ 0.554\end{tabular} \\ \hline
\textbf{Surprised} & \begin{tabular}[c]{@{}l@{}}$\mu$\\ $\sigma$\end{tabular} & \begin{tabular}[c]{@{}c@{}}0.3841\\ 0.6438\end{tabular} & \begin{tabular}[c]{@{}c@{}}0.2444\\ 0.4862\end{tabular} & \begin{tabular}[c]{@{}c@{}}0.2768\\ 0.5038\end{tabular} & \begin{tabular}[c]{@{}c@{}}0.6885\\ 1.211\end{tabular} & \begin{tabular}[c]{@{}c@{}}0.3588\\ 0.6817\end{tabular} & \begin{tabular}[c]{@{}c@{}}0.3832\\ 0.7165\end{tabular} \\ \hline

    \end{tabular}
    \caption{Mean and standard deviation of the emotion intensity scores estimated by the OpenFace~\cite{8373812} analysis tool for each emotion and intensity level. Results are reported for the Real dataset, the original CVAE-generated expressions and the Enhanced expressions. Higher mean values indicate stronger expression intensity, while the standard deviation reflects the variability across samples.}
    \label{tab:domain_dist}
\end{table}

\section{Final Considerations}

In this work, we proposed a method for generating controllable facial expressions in virtual humans using Conditional Variational Autoencoders trained on real facial expression data. By modeling six basic emotions at two intensity levels, our approach enables the synthesis of expressive facial behaviors without requiring actor-specific supervision, motion-capture transfer, or manual artistic authoring. The proposed framework demonstrates that meaningful emotional representations can be learned from a relatively small dataset comprising only 7,680 facial expression samples extracted from real performances while maintaining controllability across emotion categories and intensity levels.

The results indicate that the CVAE successfully captures the overall structure of facial expressions and generates coherent emotional variations. Furthermore, the proposed residual enhancement framework effectively compensates for the smoothing effects commonly associated with variational generative models by recovering expressive facial dynamics through temporal residual learning. Statistical comparisons between real, generated, and enhanced data showed a consistent increase in correspondence after the enhancement stage, suggesting that the residual model acts as a refinement mechanism rather than introducing entirely new expression patterns. Additional evaluation using Py-Feat~\cite{cheong2023pyfeatpythonfacialexpression} based emotion recognition further supported these findings. While the original CVAE-generated expressions occasionally exhibited reduced emotion recognizability, the enhanced expressions generally recovered or improved recognition performance across most emotion categories. These results suggest that the residual enhancement stage not only improves the statistical correspondence between real and synthetic facial controls but also helps preserve the emotional characteristics perceived by an independent facial analysis system.

Overall, the findings suggest that CVAEs constitute a promising framework for modeling and synthesizing emotional intensity variations in affective computing and character animation contexts. In future work, 
perceptual studies with human participants will be conducted to evaluate the realism, naturalness, emotional intensity, and recognizability of the generated expressions, providing a complementary user-centered assessment to the objective measures employed in this study. In addition to the statistical evaluation using formal equivalence methods, we also plan to compare the CVAE with other state-of-the-art facial expression generation methods.


\section*{Acknowledgment}

This result was achieved in cooperation with HP Brasil Indústria e Comércio de Equipamentos Eletrônicos LTDA, using incentives of Brazilian Informatics Law (Law nº 8.2.48 of 1991). Also this work was supported by Kunumi Institute. The authors thank the institution for its financial support and commitment to advancing scientific research.




\bibliographystyle{IEEEtran}
\bibliography{example}

@inproceedings{Blanz1999,
author = {Blanz, Volker and Vetter, Thomas},
title = {A morphable model for the synthesis of 3D faces},
year = {1999},
isbn = {0201485605},
publisher = {ACM Press/Addison-Wesley Publishing Co.},
address = {USA},
url = {https://doi.org/10.1145/311535.311556},
doi = {10.1145/311535.311556},
booktitle = {Proceedings of the 26th Annual Conference on Computer Graphics and Interactive Techniques},
pages = {187–194},
numpages = {8},
series = {SIGGRAPH '99}
}

@inproceedings{Lance1990,
author = {Williams, Lance},
title = {Performance-driven facial animation},
year = {1990},
isbn = {0897913442},
publisher = {Association for Computing Machinery},
address = {New York, NY, USA},
url = {https://doi.org/10.1145/97879.97906},
doi = {10.1145/97879.97906},
booktitle = {Proceedings of the 17th Annual Conference on Computer Graphics and Interactive Techniques},
pages = {235–242},
numpages = {8},
location = {Dallas, TX, USA},
series = {SIGGRAPH '90}
}

@article{YAN2025112899,
title = {Synthetic data for enhanced privacy: A VAE-GAN approach against membership inference attacks},
journal = {Knowledge-Based Systems},
volume = {309},
pages = {112899},
year = {2025},
issn = {0950-7051},
doi = {https://doi.org/10.1016/j.knosys.2024.112899},
url = {https://www.sciencedirect.com/science/article/pii/S0950705124015338},
author = {Jian’en Yan and Haihui Huang and Kairan Yang and Haiyan Xu and Yanling Li}
}

@article{ISLAM2021105950,
title = {Crash data augmentation using variational autoencoder},
journal = {Accident Analysis and Prevention},
volume = {151},
pages = {105950},
year = {2021},
issn = {0001-4575},
doi = {https://doi.org/10.1016/j.aap.2020.105950},
url = {https://www.sciencedirect.com/science/article/pii/S000145752031770X},
author = {Zubayer Islam and Mohamed Abdel-Aty and Qing Cai and Jinghui Yuan}
}

@article{Bai:2022,
author = {Bai, Wenjun and Quan, Changqin and Luo, Zhi-Wei},
year = {2022},
month = {01},
pages = {},
title = {Data-driven Dimensional Expression Generation via Encapsulated Variational Auto-Encoders},
volume = {15},
journal = {Cognitive Computation},
doi = {10.1007/s12559-021-09973-z}
}

@inproceedings{10.1145/3680528.3687669,
author = {Qiu, Feng and Zhang, Wei and Liu, Chen and An, Rudong and Li, Lincheng and Ding, Yu and Fan, Changjie and Hu, Zhipeng and Yu, Xin},
title = {FreeAvatar: Robust 3D Facial Animation Transfer by Learning an Expression Foundation Model},
year = {2024},
isbn = {9798400711312},
publisher = {Association for Computing Machinery},
address = {New York, NY, USA},
url = {https://doi.org/10.1145/3680528.3687669},
doi = {10.1145/3680528.3687669},
booktitle = {SIGGRAPH Asia 2024 Conference Papers},
articleno = {42},
numpages = {11},
location = {Tokyo, Japan},
series = {SA '24}
}

@inproceedings{kaisiyuan2020mead,
 author = {Wang, Kaisiyuan and Wu, Qianyi and Song, Linsen and Yang, Zhuoqian and Wu, Wayne and Qian, Chen and He, Ran and Qiao, Yu and Loy, Chen Change},
 title = {MEAD: A Large-scale Audio-visual Dataset for Emotional Talking-face Generation},
 booktitle = {ECCV},
 month = Augest,
 year = {2020}
}

@inproceedings{NIPS2015_8d55a249,
 author = {Sohn, Kihyuk and Lee, Honglak and Yan, Xinchen},
 booktitle = {Advances in Neural Information Processing Systems},
 editor = {C. Cortes and N. Lawrence and D. Lee and M. Sugiyama and R. Garnett},
 pages = {},
 publisher = {Curran Associates, Inc.},
 title = {Learning Structured Output Representation using Deep Conditional Generative Models},
 url = {https://proceedings.neurips.cc/paper_files/paper/2015/file/8d55a249e6baa5c06772297520da2051-Paper.pdf},
 volume = {28},
 year = {2015}
}

@dataset{livingstone_2018_1188976,
  author       = {Livingstone, Steven R. and
                  Russo, Frank A.},
  title        = {The Ryerson Audio-Visual Database of Emotional
                   Speech and Song (RAVDESS)
                  },
  month        = apr,
  year         = 2018,
  publisher    = {Zenodo},
  version      = {1.0.0},
  doi          = {10.5281/zenodo.1188976},
  url          = {https://doi.org/10.5281/zenodo.1188976},
}

@misc{he2015deepresiduallearningimage,
      title={Deep Residual Learning for Image Recognition}, 
      author={Kaiming He and Xiangyu Zhang and Shaoqing Ren and Jian Sun},
      year={2015},
      eprint={1512.03385},
      archivePrefix={arXiv},
      primaryClass={cs.CV},
      url={https://arxiv.org/abs/1512.03385}, 
}

@misc{cheong2023pyfeatpythonfacialexpression,
      title={Py-Feat: Python Facial Expression Analysis Toolbox}, 
      author={Jin Hyun Cheong and Eshin Jolly and Tiankang Xie and Sophie Byrne and Matthew Kenney and Luke J. Chang},
      year={2023},
      eprint={2104.03509},
      archivePrefix={arXiv},
      primaryClass={cs.CV},
      url={https://arxiv.org/abs/2104.03509}, 
}

@INPROCEEDINGS{8373812,
  author={Baltrusaitis, Tadas and Zadeh, Amir and Lim, Yao Chong and Morency, Louis-Philippe},
  booktitle={2018 13th IEEE International Conference on Automatic Face and Gesture Recognition (FG 2018)}, 
  title={OpenFace 2.0: Facial Behavior Analysis Toolkit}, 
  year={2018},
  volume={},
  number={},
  pages={59-66},
  doi={10.1109/FG.2018.00019}}
%\begin{thebibliography}{1}

%\end{thebibliography}

\end{document}